\documentclass[10pt,twocolumn,letterpaper]{article}

\usepackage{cvpr}              

\usepackage{graphicx}
\usepackage{amsmath}
\usepackage{amssymb}
\usepackage{booktabs}
\usepackage{placeins}
\usepackage{wrapfig}
\usepackage[export]{adjustbox}
\usepackage{enumitem}
\usepackage[accsupp]{axessibility}

\usepackage[pagebackref,breaklinks,colorlinks]{hyperref}

\usepackage[capitalize]{cleveref}
\crefname{section}{Sec.}{Secs.}
\Crefname{section}{Section}{Sections}
\Crefname{table}{Table}{Tables}
\crefname{table}{Tab.}{Tabs.}

\def\cvprPaperID{7748} 
\def\confName{CVPR}
\def\confYear{2022}

\begin{document}

\title{Generating Representative Samples for  Few-Shot Classification}

\author{Jingyi Xu\\
Stony Brook University\\
{\tt\small jingyixu@cs.stonybrook.edu}
\and
Hieu Le\thanks{Work done outside of Amazon}\\
Amazon Robotics\\
{\tt\small ahieu@amazon.com}
}
\maketitle

\begin{abstract}

Few-shot learning (FSL) aims to learn new categories with a few visual samples per class. Few-shot class representations are often biased due to data scarcity.  To mitigate this issue, we propose to generate visual samples based on semantic embeddings using a conditional variational autoencoder (CVAE) model. We train this CVAE model on base classes and use it to generate features for novel classes. More importantly, we guide this VAE to strictly generate representative samples by removing non-representative samples from the base training set when training the CVAE model. We show that this training scheme enhances the representativeness of the generated samples and therefore, improves the few-shot classification results. Experimental results show that our method improves three FSL baseline methods by substantial margins, achieving state-of-the-art few-shot classification performance on miniImageNet and tieredImageNet datasets for both 1-shot and 5-shot settings. Code is available at:
\url{https://github.com/cvlab-stonybrook/fsl-rsvae}.
\end{abstract}

\section{Introduction}
\label{sec:intro}

Few-shot learning (FSL) methods aim to learn useful representations with limited training data. They are extremely useful for situations where machine learning solutions are required but large labelled datasets are not trivial to obtain (e.g. rare medical conditions \cite{Ouyang2020SelfSupervisionWS, Wang2021FewShotLB}, rare animal species \cite{cub}, failure cases in autonomous systems \cite{Rezaei2020ZeroshotLA,Majee2021FewShotLF,Majee2021MetaGM}).
Generally, FSL methods learn knowledge from a fixed set of base classes with a surplus of labelled data and then adapt the learned model to a set of novel classes for which only a few training examples are available \cite{Wang2019GeneralizingFA}. 

\definecolor{azure(colorwheel)}{rgb}{0.0, 0.5, 1.0}
\definecolor{amber}{rgb}{1.0, 0.75, 0.0}

\begin{figure}[t]
\begin{center}
\includegraphics[width=\linewidth]{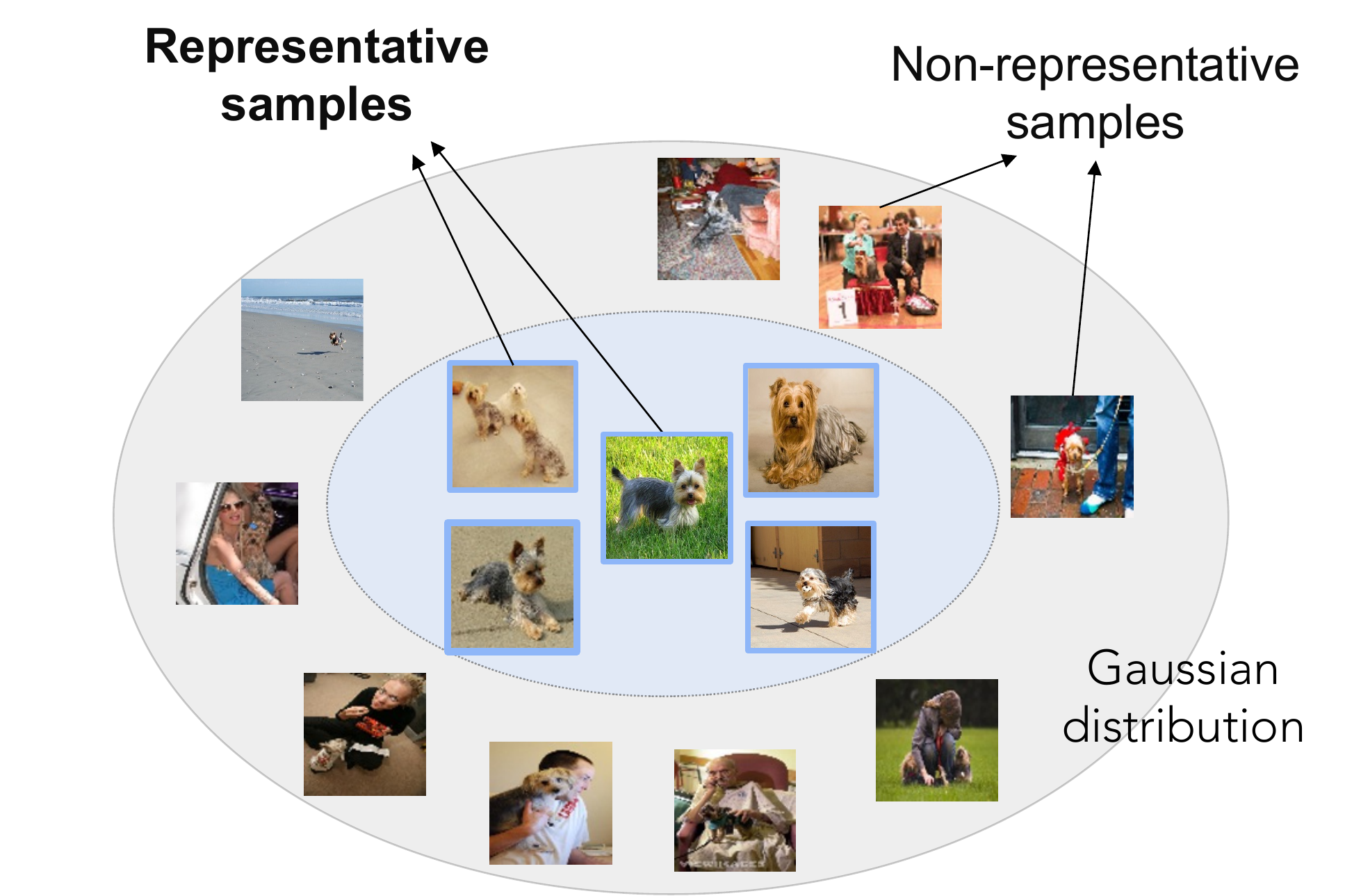}
\end{center}
\caption{
\textbf{Representative Samples.} 
We refer representative samples to the “easy-to-recognize” samples that faithfully reflect the key characteristics  of  the  category.  
We identify those samples and then use them to train a VAE model for feature generation, conditioned on class-representative semantic embeddings.
We show that the generated data significantly improves few-shot classification performance.
}
\label{fig:teaser}
\end{figure}


Many FSL methods \cite{protonet,proto_vae,zhang2021prototype,metabaseline,protonet,Liu2020AnEO,Xing2019AdaptiveCF_AM3} employ a prototype-based classifier for its simplicity and good performance. They aim to find a prototype for each novel class such that it is close to the testing samples of the same class and far away from testing samples for other classes. 
However, it is challenging to estimate a representative prototype just from a few available support samples \cite{Yang2021FreeLF,Liu2020PrototypeRF}. 
An effective strategy to enhance the representativeness of the prototype is to employ textual semantic embeddings learned via NLP models\cite{Radford2021LearningTV,Pennington2014GloVeGV,Miller1992WordNetAL,Devlin2019BERTPO} using large unsupervised text corpora \cite{zhang2021prototype,Xing2019AdaptiveCF_AM3}. These semantic embeddings implicitly associate a class name, such as ``Yorkshire Terriers'', with the class representative semantic attributes such as “smallest dog” or “long coat” \cite{terns} ( Fig. \ref{fig:teaser}), providing strong and unbiased priors for category recognition.  

For the most part, current FSL methods 
focus on learning to adaptively leverage the semantic information to complete the original biased prototype estimated from the few available samples. For example, the recent FSL method of Zhang \etal \cite{zhang2021prototype} 
 learns to fuse the primitive knowledge and attribute features into a representative prototype, depending on the set of given few-shot samples. Similarly, Xing \etal \cite{Xing2019AdaptiveCF_AM3} propose a method that computes an adaptive mixture coefficient to combine features from the visual and textual modalities. 
 However, learning to recover an arbitrarily biased prototype is challenging due to the drastic variety of the possible combinations of few-shot samples.

In this paper, we propose a novel FSL method to obtain class-representative prototypes.
Inspired by zero-shot learning (ZSL) methods\cite{Guo2021ANP,Ba2015PredictingDZ,Zhang2017LearningAD}, we propose to generate visual features via a variational autoencoder (VAE) model \cite{CVAE} conditioned on the semantic embedding of each class.  
This VAE model learns to associate a distribution of features to a conditioned semantic code. We assume that such association generalizes across the base and novel classes \cite{Mishra2018AGM,Arora2018GeneralizedZL}. Therefore, the model trained with sufficient data from the base classes can generate novel-class features that align with the real unseen features. We then use the generated features together with the few-shot samples to construct class prototypes. We show that this strategy achieves state-of-the-art results on both \textit{mini}ImageNet and \textit{tiered}ImageNet datasets. 
It works exceptionally well for 1-shot scenarios where our method outperforms state-of-the-art methods\cite{Wertheimer2021FewShotCW,Ye2020FewShotLV}  by $5\sim6\%$ in terms of classification accuracy.

Moreover, to enhance the representativeness of the prototype, we guide the VAE to generate more \textit{representative} samples. 
Here we refer representative samples to the ``\textit{easy-to-recognize}'' samples that faithfully reflect the key characteristics of the category (see Fig. \ref{fig:teaser}). The embeddings of these representative samples often lie close to their corresponding class centers, which are particularly useful for constructing class-representative prototypes.

Specifically, we guide the VAE model to generate representative samples by selecting only representative data from the base classes for training it. In essence, our VAE model is trained to model the data distribution of the training set. As the training set contains only representative data, the trained VAE model outputs samples that are also representative. Specifically, to select those representative features, we first assume that the feature vectors of each class follow a multivariate Gaussian distribution and estimate this distribution for each base class. Based on these distributions, we compute the probability of each sample belonging to its corresponding category to measure the representativeness for the sample. We filter out the non-representative samples and train the VAE using only representative samples. Interestingly, we show that the representativeness of the training set highly corresponds to the accuracy of the few-shot classifier. We obtain the highest accuracy when training the VAE with the most representative samples. In this case, we only use a small percentage of the whole training set, e.g., 10\% for the case of \textit{mini}Imagenet dataset, to obtain the best results. Our analyses show that 
this approach consistently improves the FSL classification performance by $1\sim2\%$ across all benchmarks for three different baselines\cite{metabaseline,protonet,Liu2020AnEO}.

Our main contributions can be summarized as follows: 
\begin{itemize}
\item We are the first to use a VAE-based feature generation approach conditioned on class semantic embeddings for few-shot classification. 
\item We propose a novel sample selection method to collect representative samples. We use these samples to train a VAE model to obtain reliable data points for constructing class-representative prototypes.
\item Our experiments show that our methods achieve state-of-the-art performance on two challenging datasets, \textit{tiered}ImageNet and \textit{mini}ImageNet. 
\end{itemize}

We summarize related FSL works in Section \ref{sec:rw}. Section \ref{sec:method} provides a rundown of our approach. Section \ref{sec:exp} reports the main results obtained with our method. In section \ref{sec:ana}, we provide multiple analyses to clarify different aspects of our methods.

%
%


\section{Related Work}

\begin{figure*}[!ht]
\begin{center}
\includegraphics[width=1.9\columnwidth]{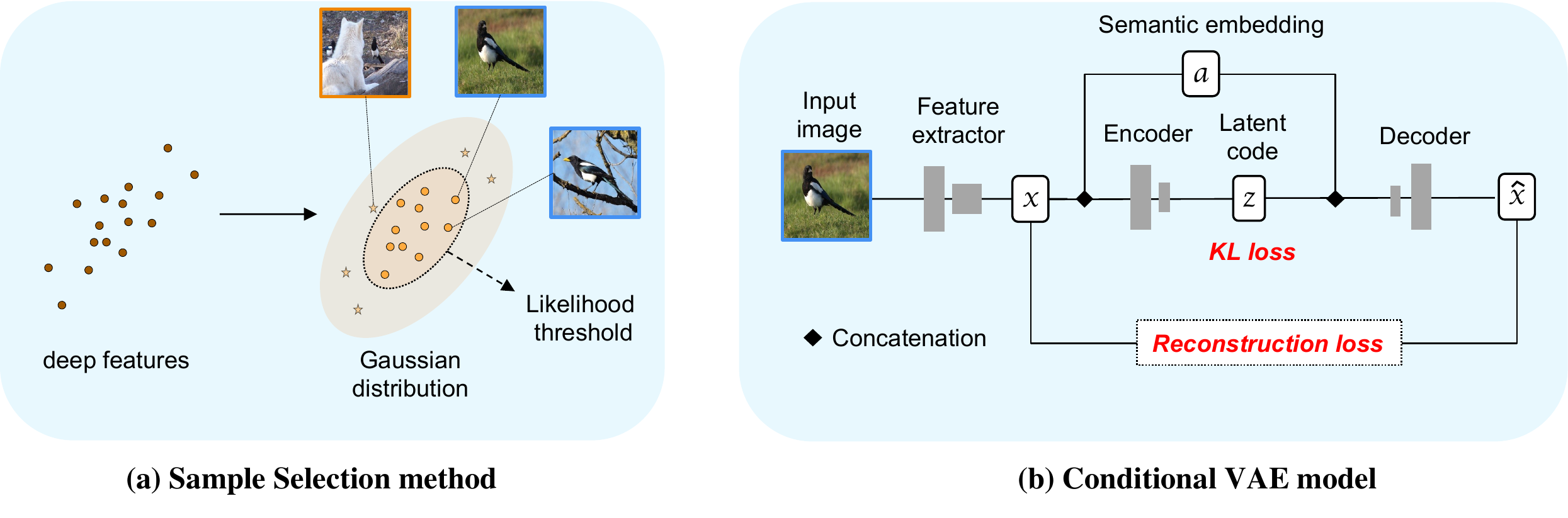}
\caption{\textbf{Overview -- }
The key aspect of our approach is to subset our training set to the most representative samples to train a conditional VAE model that generates more representative features. \textbf{(a)} To select representative samples, we assume that the features of each class follow a multivariate Gaussian distribution. We estimate the distribution parameters and compute a probability for each data point belonging to the class distribution. We identify a set of representative samples by setting a threshold on the probability.  \textbf{(b)} We train a VAE to generate visual features, conditioned on the semantic embedding of each class. Using only representative samples (the output of the sample selection step) to train this VAE model improves the representativeness of the generated samples.
}
\label{fig:overview2}
\end{center}

\end{figure*}


\label{sec:rw}

\textbf{Few-shot Learning.} 
FSL is helpful when we only have limited labeled training data \cite{le2020physicsbased,Le_2020_ECCV,Le-etal-ICCV19,Le2016GeodesicDH,Borowicz2019AerialtrainedDL,LeICCV2017,m_Le-etal-ECCV18}. 
Representative FSL approaches include metric learning based \cite{matchingnet,protonet,relationnet,Ye2020FewShotLV,Yang2021FreeLF,Tian2020RethinkingFI,Zhang2020DeepEMDFI}, optimization based \cite{maml,metasgd,meta-opt,Lifchitz2019DenseCA,Santoro2016MetaLearningWM,Rajeswaran2019MetaLearningWI,Scott2018AdaptedDE,Liu2020PrototypeRF}, and data augmentation based methods \cite{dagan,delta-encoder,wang2018lowshot,JingyiICCV21}. 
Similar to our method, some FSL methods use semantic information to improve the few-shot classifiers \cite{zhang2021prototype,Xing2019AdaptiveCF_AM3,Tokmakov2019LearningCR,Peng2019FewShotIR,Hu2019WeaklysupervisedCF}.  Zhang \etal \cite{zhang2021prototype} and Xing \etal\cite{Xing2019AdaptiveCF_AM3} propose methods that learn to adaptively combine the visual features and the semantic features to obtain an unified cross-modality representation for each class. These two methods focus on the fusing strategies that combine features of the two domains. 
Hu \etal\cite{Hu2019WeaklysupervisedCF} propose to disentangle the visual features into the sub-spaces that associate to different semantic attributes. The FSL method of Peng \etal \cite{Peng2019FewShotIR} uses semantic information to infer a classifier for novel classes and adaptively combines this classifier with the few-shot samples. Our method is the first FSL method that uses a conditional VAE model to directly generate {visual} features, conditioned on the semantic embedding of each class. 

\textbf{Conditional Variational Autoencoder.} The practice of using a conditional VAE to model a feature distribution has been used before in many computer vision tasks such as image classification \cite{variational_fewshot,aligned_vae,proto_vae,JingyiICCV21}, image generation \cite{Esser2018AVU,Liu2017UnsupervisedIT}, image restoration \cite{Du2020ConditionalVI}, or video processing \cite{Pan2019VideoGF}. Using VAE models for generating features conditioned on the corresponding semantic embedding is fairly common in ZSL methods\cite{Guo2021ANP,Ba2015PredictingDZ,Zhang2017LearningAD,aligned_vae,Mishra2018AGM,Yu2020EpisodeBasedPG}. Mishra \etal\cite{Mishra2018AGM} are the first to propose to use a conditional VAE for ZSL where they view ZSL as a case of missing data. They find that such an approach can handle well the domain shift problem. Similarly, Arora \etal\cite{Arora2018GeneralizedZL} show that a conditional VAE can be used together with a GAN system to synthesize images for unseen classes effectively. Keshari \etal \cite{Keshari2020GeneralizedZL} focus on generating a specific set of \textit{hard} samples which are closer to another class and the decision boundary. For the most part, ZSL methods aim to model the whole distribution of data \cite{aligned_vae,Ma2020AVA,Bendre2021GeneralizedZL,Gu2020GeneralizedZL}, while our method focuses on modeling the distribution of representative samples useful for constructing the class-representative prototypes.

\textbf{Sample Selection.} To the best of our knowledge, we are the first to propose using a sample selection method for selecting training samples for a VAE model. Here we select only representative samples for training the VAE. 
This is a new sample selection regime since mainstream sample selection works  mainly focus on identifying the most informative samples \cite{Bappy2017TheIO,Le_2019_CVPR_Workshops} for training their models, which is widely used in active-learning\cite{Settles2009ActiveLL,Li2013AdaptiveAL}. In FSL, Chang \etal\cite{Chang2021OnTI} propose a method to select the most informative data that should be annotated for a few-shot text generation system. Zhou \etal\cite{Zhou2020LearningTS} propose a method to select the useful \textit{base classes} to train their model, while our work selects useful individual samples within an arbitrary set of base classes.

\section{Method}
\label{sec:method}


\subsection{Problem Definition} 
\label{sec:definition}
In a typical few-shot classification setting, we are given a set of data-label pairs D = $\{(x^i, y^i)\}$. Here $x^i \in R^d$ is the feature vector of a sample and $y^i \in C$, where $C$ denotes the set of classes. The set of classes is divided into base classes $C_b$ and novel classes $C_n$. The sets of class $C_b$ and $C_n$ are disjoint, $i.e.$ $C_b \cap C_n = \emptyset$. 
For a $N$-way $K$-shot problem, we sample $N$ classes from the novel set $C_n$, and $K$ samples are available for each class. $K$ is often small ($i.e.$, $K = 1$ or $K = 5$). Our goal is to classify query samples correctly using the few samples from the support set.

\subsection{Overall Pipeline} 
\label{sec:overall_pipeline}

Fig. \ref{fig:overview2} gives an overview of our sample selection method and VAE training approach. We propose a method to select a set of representative samples from a set of base classes.
We use these selected representative data to train a conditional VAE model for feature generation. To select representative samples, we assume that the features of each class follow a multivariate Gaussian distribution. We estimate the parameters for each class distribution and compute the probability for each data point belonging to its class. By setting a threshold on the probabilities, we identify a set of representative samples. We then use these selected representative samples to train a VAE model that generates samples conditioned on the semantic attributes of each class. 


We train this VAE on the base classes and use the trained model to generate samples for the novel classes. The generated features are then used together with the few-shot samples to construct the prototype for each class. Our method is a simple plug-and-play module and can be built on top of any pretrained feature extractors. In our experiments, we show that our method consistently improves three baseline few-shot classification methods: Meta-Baseline \cite{metabaseline}, ProtoNet \cite{protonet} and E3BM \cite{Liu2020AnEO} by large margins. 

\subsubsection{Class-representative Sample Selection}

 In this paper, we are interested in representative samples as they can serve as reliable data points for constructing a class-representative prototype\cite{metabaseline,protonet}. The main idea is to train a feature generator with only representative data to obtain more representative generated samples.
 
 To select the representative features,
 we assume that the feature distribution of the base classes follows a Gaussian distribution and estimate the parameters of this distribution for each class. We calculate the Gaussian mean of a base class $i$ as the mean of every single dimension in the vector:
  \begin{equation}\label{eq:mean}
     \mu^i = \frac{1}{n^i}\sum_{j=1}^{n^i} x^j,
\end{equation}
 where $x^j$ is a feature vector of the $j$-th sample from the base class $i$ and $n^i$ is the total number of samples in class $i$. 
The covariance matrix $\Sigma^i$ for the distribution of class $i$ is calculated as:
 \begin{equation}\label{eq:vae}
     \Sigma^i = \frac{1}{n^i-1} \sum\limits_{j=1}^{n^i}(x^j-\mu^i)(x^j-\mu^i)^T.
\end{equation}
Once we estimate the parameters of the Gaussian distribution using the adequate samples from the base classes, the probability density of observing a single feature, $x^j$, being generated from the Gaussian distribution of class $i$ is given by:
 \begin{equation}\label{eq:likelihood}
 \begin{aligned}
     p(x^j|\mu^i, \Sigma^i) &=  \frac{\textnormal{exp}\{-\frac{1}{2}(x^j - \mu^i)^T{\Sigma^i}^{-1}(x^j - \mu^i)\}}{(2\pi)^{k/2}|\Sigma^i|^{1/2}},
  \end{aligned}
\end{equation}
where $k$ is the dimension of the feature vector.

Here we assume that the probability of a single sample belongs to its category's distribution reflects the representativeness of the sample, \textit{i.e.}, the higher the probability, the more representative the sample is. By setting a threshold $\epsilon$ on the estimated probability, we filter out those samples with small probabilities and get a set of representative features for class $i$:
 \begin{equation}\label{eq:representative}
   \mathbb{D}^i = \{x^j \text{  } | \text{  }  p(x^j|\mu^i, \Sigma^i) > \epsilon\}, 
\end{equation}
where $\mathbb{D}^i$ stores the features for class $i$ with the probabilities larger than a threshold $\epsilon$.


\subsubsection{Conditional VAE Model for Feature Generation}

We use our sample selection method to select a set of representative samples and use them for training our feature generation model. 
We develop our feature generator  based on a conditional variational autoencoder (VAE) architecture\cite{CVAE} (see Fig. \ref{fig:overview2}b). The VAE is composed of an Encoder $E(x, a)$, which maps a visual feature $x$ to a latent code $z$, and a decoder $G(z, a)$ which reconstructs $x$ from $z$. Both $E$ and $G$ are conditioned on the semantic embedding $a$.  The loss function for training the  VAE for a feature $x^j$ of class $i$ can be defined as:
 \begin{equation}\label{eq:cvae}
  \begin{aligned}
      L_{V}(x^j) = & \textnormal{KL} \left( q(z|x^j,a^i)||p(z|a^i) \right) \\ 
      & - \textnormal{log}p(x^j|z,a^i),
  \end{aligned}
\end{equation}
where $a^i$ is the semantic embedding of class $i$. The first term is the Kullback-Leibler divergence between the VAE posterior $q(z|x,a)$ and a prior distribution $p(z|a)$. The second term is the decoder's reconstruction error. $q(z|x,a)$ is modeled as $E(x, a)$ and $p(x|z,a)$ is equal to $G(z, a)$. The prior distribution is assumed to be $\mathcal{N}(0,I)$ for all classes. 

The loss for training the feature generator is the loss over all selected representative training samples:

 \begin{equation}
L_V = \sum_{i=1}^{C_b}\sum_{x \in \mathbb{D}^i} L_V(x)
\end{equation}
\newcommand{\best}[1]{{\textbf{#1}}} 
\newcommand{\secondbest}[1]{{{#1}}}

   \begin{table*}[t] 
    \centering
  \resizebox{0.83\textwidth}{!}{%
    \begin{tabular}{l|l|cc|cc}
      \hline
      Method  & Backbone  & \multicolumn{2}{c|}{\textit{mini}ImageNet} & \multicolumn{2}{c}{\textit{tiered}ImageNet}
                \\
      & & 1-shot & 5-shot & 1-shot & 5-shot  \\
      \hline
      Matching Net \cite{matchingnet}  & ResNet-12 & 65.64 $\pm$ 0.20 & 78.72 $\pm$ 0.15  & 68.50 $\pm$ 0.92 & 80.60  $\pm$ 0.71  \\
      MAML \cite{maml}& ResNet-18 & 64.06 $\pm$ 0.18 & 80.58 $\pm$ 0.12& - & -  \\
      SimpleShot \cite{Wang2019SimpleShotRN}     & ResNet-18 & 62.85  $\pm$ 0.20 & 80.02  $\pm$ 0.14 & 69.09 $\pm$ 0.22 & 84.58 $\pm$ 0.16    \\
      CAN \cite{Hou2019CrossAN}& ResNet-12 & 63.85 $\pm$ 0.48 & 79.44 $\pm$ 0.34 & 69.89 $\pm$ 0.51 & 84.23 $\pm$ 0.37   \\
      S2M2 \cite{Mangla2020ChartingTR}& ResNet-18 & 64.06 $\pm$ 0.18 & 80.58 $\pm$ 0.12& - & -  \\
      TADAM \cite{Oreshkin2018TADAMTD} & ResNet-12 & 58.50 $\pm$ 0.30 &76.70  $\pm$ 0.30 & 62.13 $\pm$ 0.31 &81.92 $\pm$ 0.30 \\
      AM3 \cite{Xing2019AdaptiveCF_AM3} & ResNet-12 & 65.30 $\pm$ 0.49 & 78.10 $\pm$ 0.36 & 69.08 $\pm$ 0.47 & 82.58 $\pm$ 0.31 \\
      DSN \cite{Simon2020AdaptiveSF} & ResNet-12 & 62.64 $\pm$ 0.66 & 78.83 $\pm$ 0.45 & 66.22 $\pm$ 0.75 & 82.79 $\pm$ 0.48 \\
      Variational FSL \cite{variational_fewshot} & ResNet-12 & 61.23 $\pm$ 0.26 & 77.69 $\pm$ 0.17 & - & - \\
      MetaOptNet \cite{meta-opt} & ResNet-12 & 62.64 $\pm$ 0.61 & 78.63 $\pm$ 0.46 & 65.99 $\pm$ 0.72 & 81.56 $\pm$ 0.53 \\
      Robust20-distill \cite{Dvornik2019DiversityWC} & ResNet-18 & 63.06 $\pm$ 0.61& 80.63 $\pm$ 0.42& 65.43 $\pm$ 0.21 &70.44 $\pm$ 0.32 \\
      FEAT \cite{Ye2020FewShotLV} & ResNet-12 & 66.78 $\pm$ 0.20& 82.05 $\pm$ 0.14 &70.80 $\pm$ 0.23& 84.79 $\pm$ 0.16 \\
      RFS \cite{Tian2020RethinkingFI} & ResNet-12 & 62.02 $\pm$ 0.63 & 79.64 $\pm$ 0.44 & 69.74 $\pm$ 0.72 & 84.41 $\pm$ 0.55 \\
      Neg-Cosine \cite{Liu2020NegativeMM} & ResNet-12 & 63.85 $\pm$ 0.81 & 81.57 $\pm$ 0.56 & - &- \\
      FRN \cite{Wertheimer2021FewShotCW} & ResNet-12 & 66.45 $\pm$ 0.19 &82.83 $\pm$ 0.13 &71.16 $\pm$ 0.22 & {86.01 $\pm$ 0.15} \\
      \hline
      Meta-Baseline \cite{metabaseline} &ResNet-12& 63.17 $\pm$ 0.23 & 79.26 $\pm$ 0.17 & 68.62 $\pm$ 0.27 & 83.29 $\pm$ 0.18 \\
      Meta-Baseline + SVAE (Ours)&ResNet-12& 69.96 $\pm$ 0.21 & 79.92 $\pm$ 0.16 & 73.05 $\pm$ 0.24 & 83.96 $\pm$ 0.18 \\
      Meta-Baseline + R-SVAE (Ours) & ResNet-12 & 72.79 $\pm$ 0.19 & 80.70  $\pm$ 0.16 & 73.90 $\pm$ 0.24 &  84.17 $\pm$ 0.18 \\
      \hline     
      ProtoNet \cite{Ye2020FewShotLV} & ResNet-12 & 62.39  & 80.53   & 68.23 & 84.03  \\
      ProtoNet + SVAE (Ours)& ResNet-12 & {73.01 $\pm$ 0.24}  & \secondbest{83.13 $\pm$ 0.40}  & {76.36 $\pm$ 0.65} & 85.65 $\pm$ 0.50\\
      ProtoNet + R-SVAE(Ours)& ResNet-12 & \best{74.84 $\pm$ 0.23}  & \best{83.28 $\pm$ 0.40}  & {76.98 $\pm$ 0.65} & {85.77 $\pm$ 0.50}  \\
      \hline     
      E3BM \cite{Liu2020AnEO}& ResNet-12 & 64.09 $\pm$ 0.37  & 80.29 $\pm$ 0.25   & 71.34 $\pm$ 0.41 & 85.82 $\pm$ 0.29  \\
      E3BM + SVAE (Ours) & ResNet-12 & {73.07 $\pm$ 0.39}  & {80.82 $\pm$ 0.31}  & \secondbest{79.85 $\pm$ 0.43} & \secondbest{86.82 $\pm$ 0.32}\\
      E3BM + R-SVAE(Ours)& ResNet-12 & \secondbest{73.35 $\pm$ 0.37}  & {80.95 $\pm$ 0.31}  & \best{80.46 $\pm$ 0.43} & \best{86.99 $\pm$ 0.32}  \\
      \hline
    \end{tabular}}  
    \caption{\textbf{Comparison to prior works on \textit{mini}ImageNet and \textit{tiered}ImageNet}. Average 5-way 1-shot and 5-way 5-shot accuracy (\%) with 95\% confidence intervals.
     SVAE denotes our method using the VAE trained with all features in the base set. R-SVAE denotes the one trained with only representative features. The \best{best} performance is highlighted in bold.
    }
    \label{tab:mini_tiered}%
    \vspace{-10pt}
  \end{table*}
  
\subsubsection{Constructing Class Prototypes}
After the VAE is trained on the base set, we generate a set of features for a class $y$ by inputting the respective semantic vector $a^y$ and a noise vector $z$ to the decoder $G$:
 \begin{equation}
\mathbb{G}^y = \{ \hat{x} | \hat{x} = G(z, a^y), z \sim \mathcal{N}(0, I)\}.
\end{equation}
The generated features along with the original support set features for a few-shot task is then served as the training data for a task-specific classifier. Following our baseline methods, we compute the prototype for each class and apply the nearest neighbour classifier. 
Specifically, we first compute two separated prototypes: one using the support features and the other using the generated features. Each prototype
is the mean vector of the features of each group. We then take a weighted sum of the two prototypes to obtain the final prototype $\textnormal{p}^y$  for class $y$:
 \begin{equation}
 \label{eq:combine}
 \textnormal{p}^y = w_g * \frac{1}{|\mathbb{G}^y|} {\sum}_{\hat{x}^j \in \mathbb{G}^y}
{\hat{x}^j} + w_s*\frac{1}{|\mathbb{S}^y|} {\sum}_{{x}^j \in \mathbb{S}^y}
{{x^j}}, 
\end{equation}
where $\mathbb{S}^y$ is the support set features and $(w_g,w_s)$ are the coefficients of the generated feature prototype and the real feature prototype, respectively. We classify samples by finding the nearest class prototype for an embedding query feature. 
We conduct further analysis to show that our generated features can benefit all types of classifiers (see Section \ref{sec: classifier}). Compared to the methods that correct the original biased prototype, our model 
does not require any carefully designed combination scheme.    

\section{Experiments}
\label{sec:exp}

\subsection{Experimental Settings}
\textbf{Datasets.}
We evaluate our method on two widely-used benchmarks for few-shot learning, \textit{mini}ImageNet \cite{Ravi2017OptimizationAA} and \textit{tiered}ImageNet \cite{Ren2018MetaLearningFS}. \textbf{\textit{mini}ImageNet} is a subset of the ILSVRC-12 dataset \cite{Deng2009ImageNetAL}. It contains 100 classes and each class consists of 600 images. The size of each image is 84 $\times$ 84. Following the evaluation protocol of \cite{meta_learn_lstm}, we split the 100 classes into 64 base classes, 16 validation classes, and 20 novel classes for pre-training, validation, and testing. \textbf{\textit{tiered}ImageNet} is a larger subset of ILSVRC-12 dataset, which contains 608 classes sampled from hierarchical category structure. The average number of images in each class is 1281. It is first partitioned into 34 super-categories that are split into 20 classes for training, 6 classes for validation, and 8 classes for testing. This leads to 351 actual categories for training, 97 for validation, and 160 for testing. 

\textbf{Baseline methods.} Our method can be used as a simple plug-and-play module for many existing few-shot learning methods without fine-tuning their feature extractors. We investigate three baseline few-shot classification methods used in conjunction with our method: {ProtoNet} \cite{Ye2020FewShotLV}, {Meta-Baseline} \cite{metabaseline} and E3BM \cite{Liu2020AnEO}.
%
ProtoNet is known as a strong and classic prototypical approach. In our experiments, we use the ProtoNet implementation of Ye \etal \cite{Ye2020FewShotLV}. {Meta-Baseline} \cite{metabaseline} uses a ProtoNet model to fine-tune a generic classifier via meta-learning.  E3BM \cite{Liu2020AnEO} meta-learns the ensemble of epoch-wise models to achieve robust predictions for FSL. For each baseline method, we 
extract the corresponding feature representations to train our feature generation VAE model. We then use the trained VAE to generate features and obtain the class prototypes for few-shot classification.

\textbf{Evaluation protocol.} We use the top-1 accuracy as the evaluation metric to measure the performance of our method. 
We report the accuracy on standard
5-way 1-shot and 5-shot settings with 15 query samples per class. We randomly sample 2000 episodes from the test set and report the mean accuracy with the $95\%$ confidence interval.

 \begin{figure*}[th]
  \centering
\includegraphics[width=1.\linewidth]{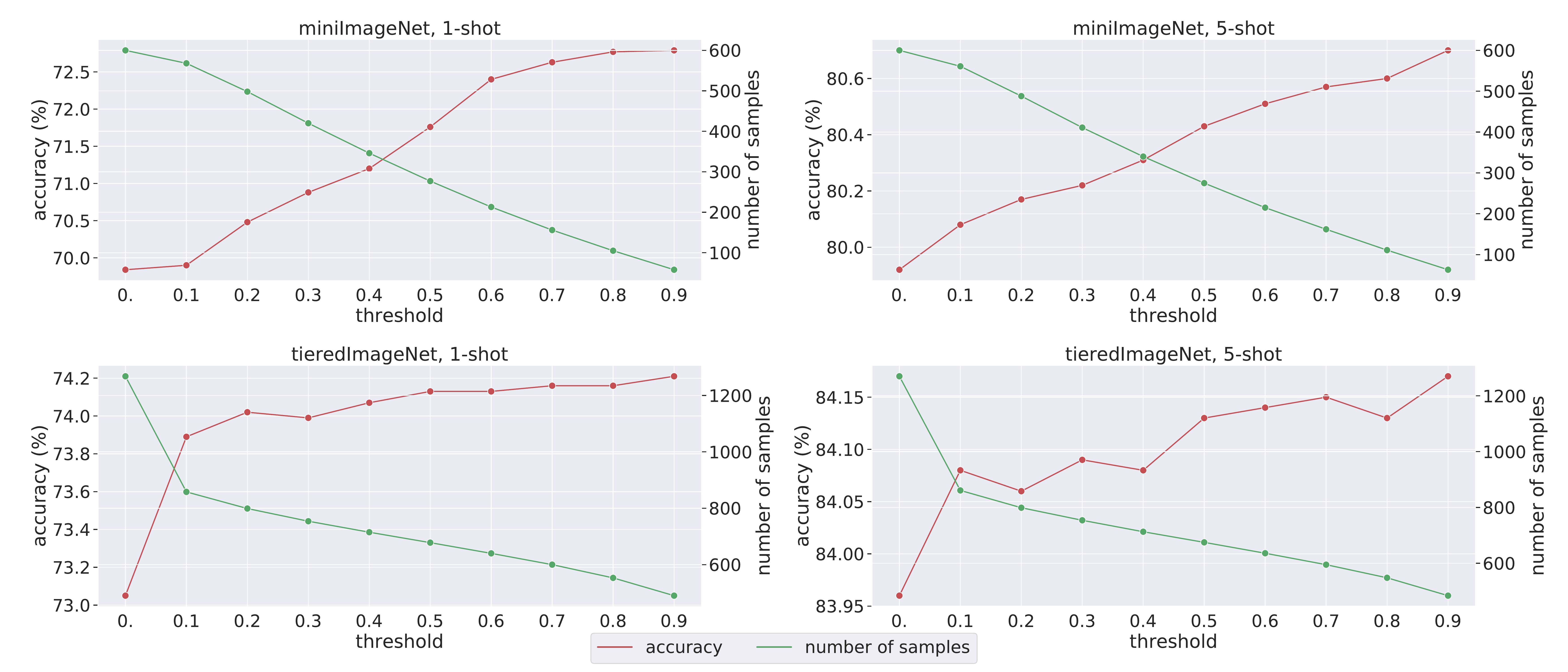}
  \caption{\textbf{Few-shot classification results with different probability thresholds.} We report the classification accuracy (\%) (red) and the number of samples (green) when setting different thresholds for the probabilities. A higher threshold means we select samples that are more representative, resulting in a less amount of training data points. In general, the classification performance increases when the number of training samples decreases with increasing representativeness thresholds. }
  \label{fig:thres}
\end{figure*}

\subsection{Implementation Details}
All the three baselines use ResNet12 backbone
as the feature extractor. The feature representation is extracted by average pooling the final residual block outputs. The dimension of the feature representation is 640 for ProtoNet \cite{Ye2020FewShotLV}, 512 for Meta-Baseline \cite{metabaseline}, and 640 for E3BM\cite{Liu2020AnEO}. For our feature generation model, both the encoder and the decoder are two-layer fully-connected (FC) networks with 4096 hidden units. LeakyReLU  and ReLU~\cite{LReLu} are the nonlinear activation functions in the hidden and output layers, respectively. The dimensions of the latent space and the semantic vector are both set to be 512. The network is trained using the Adam optimizer with $10^{-4}$ learning rate. Our semantic embeddings are extracted from CLIP \cite{Radford2021LearningTV}. We empirically set the combination weights $[ w_g, w_s]$ in Equation \ref{eq:combine} to $[\tfrac{1}{2},\tfrac{1}{2}]$ for 1-shot settings and to $[\tfrac{1}{6},\tfrac{5}{6}]$ for 5-shot settings. We set the probability threshold to 0.9 for the main experiments and discuss the performance under different values of this threshold in Section \ref{sec:threshold}.

\subsection{Results}
Table \ref{tab:mini_tiered} presents the 5-way 1-shot and 5-way 5-shot classification results of our methods on \textit{mini}ImageNet and \textit{tiered}ImageNet in comparision with previous FSL methods. Here all methods use ResNet12/ResNet18 architectures as feature extractors with input images of size 84 $\times$ 84. Thus, the comparison is fair. For the rest of the paper, we denote our VAE trained with all data as \textbf{SVAE} (\underline{S}emantic-\underline{VAE}) and the model trained with only representative data as \textbf{R-SVAE} (\underline{R}epresentative-\underline{SVAE}).

We apply our methods on top of the Meta-Baseline \cite{metabaseline}, ProtoNet\cite{Ye2020FewShotLV}, and E3BM\cite{Liu2020AnEO}. Our methods consistently improve all three baselines under all settings and for all datasets. They work particularly well under the 1-shot settings, in which sample bias is a more pronounced issue. Using the model trained on all data - SVAE, we report $6.8\% \sim 10\%$ 1-shot accuracy improvements for all three baselines. Our 1-shot performance for all the baselines outperforms the state-of-the-art method \cite{Wertheimer2021FewShotCW} by large margins. In 5-shot, our method consistently brings a $0.5 \sim 2.7 \%$ performance gains to all baselines.  

Using representative samples to train our VAE model further improves the three baseline methods under all settings and for all datasets. Compared to SVAE, training on strictly representative data improves the 1-shot classification accuracy by $0.3\% \sim 2.8 \%$ and the 5-shot classification accuracy by $0.2\% \sim 0.8 \%$. R-SVAE achieves state-of-the-art few-shot classification on \textit{mini}ImageNet dataset with the ProtoNet baseline and on \textit{tiered}ImageNet dataset with the E3BM baseline.


\begin{figure*}[!t]
  \centering
\includegraphics[width=0.9\linewidth]{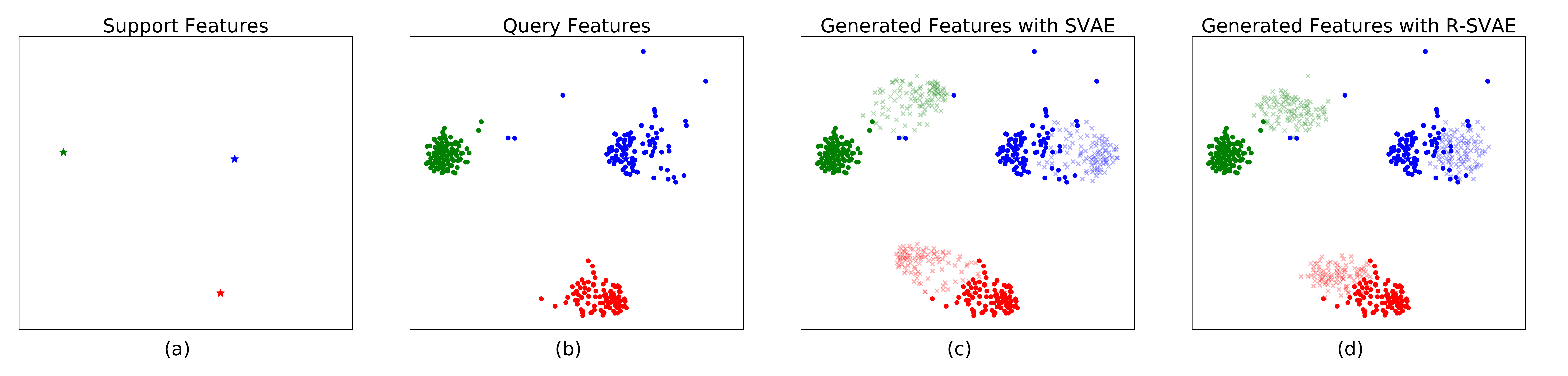}
  \\
 \caption{\textbf{Feature Visualization.} We show the t-SNE visualization of the original features (marked as dark points) and our generated features (marked as transparent points) on \textit{tiered}ImageNet dataset. Different colors represent different classes.
 From left to right, we show the original support set (a), the query set (b), the features generated by SVAE (c), and the features generated by R-SVAE (d). 
 }
   \label{fig:feats_tsne}
\end{figure*}

  \begin{figure}[]
  \centering
\includegraphics[width=1.\linewidth]{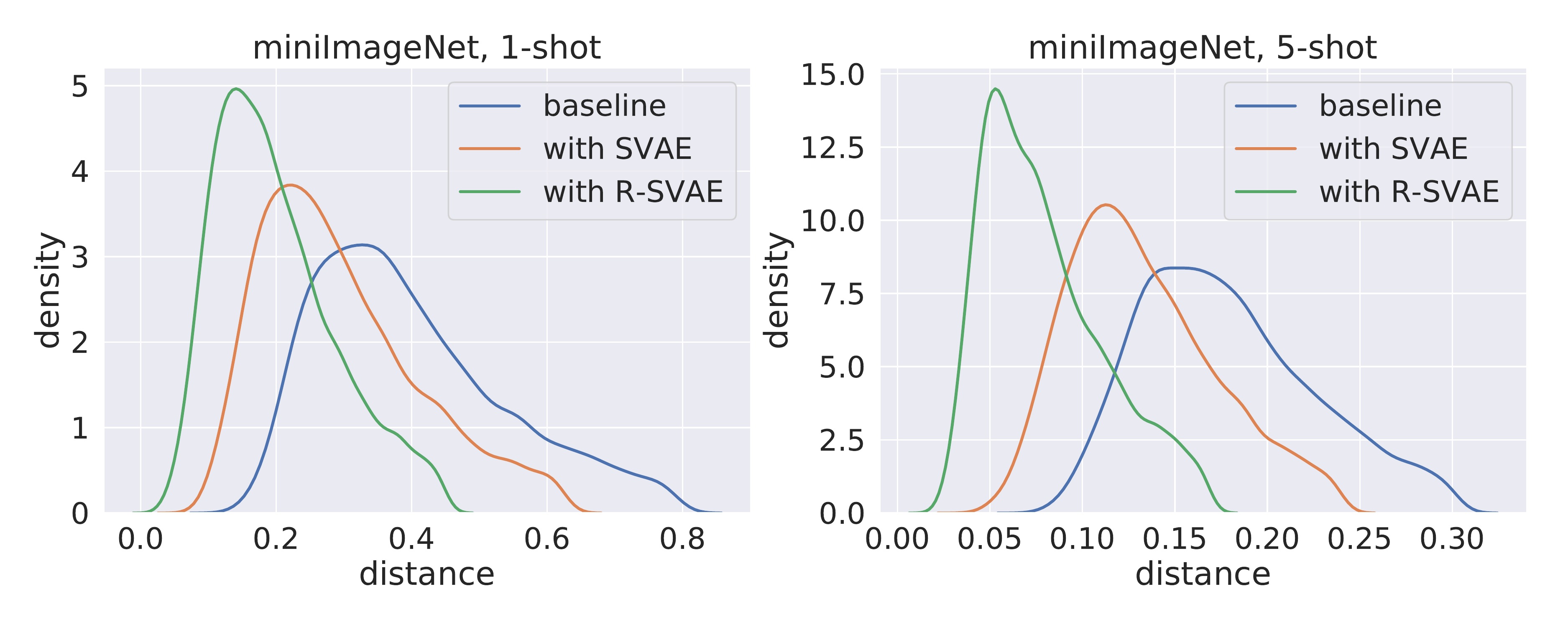}
  \caption{\textbf{Distance Distributions.} Kernel Density Estimation of the distance between the estimated prototypes and the ground truth prototype. A smaller value means the estimated prototypes are closer to the ground truth prototypes.
  }
  \label{fig:dist_density}
\end{figure}

\section{Analyses}
\label{sec:ana}
All the following analyses use the feature extractor from the Meta-Baseline method \cite{metabaseline}.

 \subsection{Analysis on the Probability Threshold}
 \label{sec:threshold}
In our main setting, we set a threshold of 0.9 on the probabilities to select those class-representative samples as the training data for our VAE model (the higher, the more representative). In this section, we conduct experiments with different threshold values to see how it affects the classifier's performance. Fig. \ref{fig:thres} shows the classification accuracy under different thresholds on \textit{mini}ImageNet and \textit{tiered}ImageNet datasets. As the threshold increases, more non-representative samples are filtered out, resulting in less training data for R-SVAE. Interestingly, we observe that the model generally performs better with higher threshold values under both 1-shot and 5-shot settings. For example, under the 1-shot setting on \textit{mini}ImageNet dataset, we only use 58 images per class on average when setting the threshold to 0.9. Training the VAE model with this small set of images improves the performance by $2.95\%$ compared with the model trained using all data in the base set with 600 images per class on average. 
The results suggest that the performance of our method strongly corresponds to the representativeness of training data. Moreover, it shows that our sample selection method provides a reliable  measurement for the representativeness of the training samples.

 \def\subboxsize{0.4\textwidth}
 \begin{figure*}[ht!]
 \centering
\hspace*{-0.1cm}\includegraphics[width=0.8\linewidth]{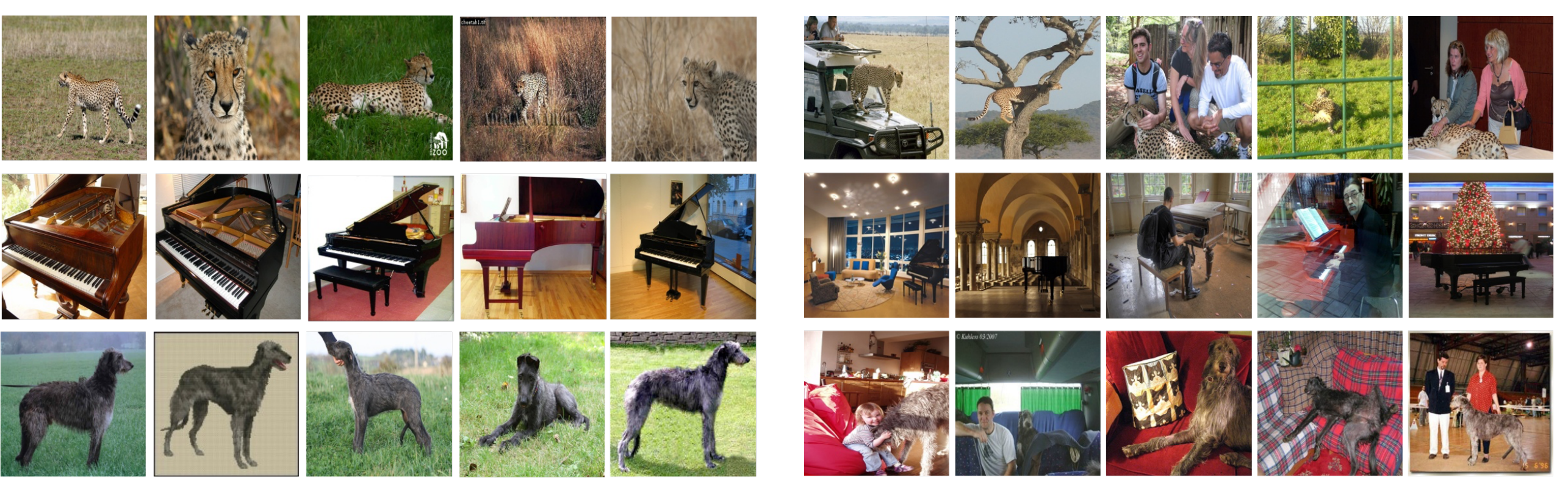}
\makebox[\subboxsize]{Representative }
\makebox[\subboxsize]{Non-representative }
 \caption{ \textbf{Examples of representative samples (left) and non-representative samples (right)}. We visualize 5 images with high probabilities and 5 images with small probabilities computed via our proposed method for 3 classes from \textit{tiered}ImageNet dataset.
    }

    \label{fig:img_visualization}
\end{figure*}

  \begin{table*}[] 
    \centering
  \resizebox{0.85\textwidth}{!}{%
    \begin{tabular}{l|ccc|ccc}
      \hline
        & \multicolumn{3}{c|}{\textit{mini}ImageNet} & \multicolumn{3}{c}{\textit{tiered}ImageNet}
           \\
          Classifier & support samples & + SVAE & + R-SVAE & support samples & + SVAE & +R-SVAE \\
      \hline
       Prototype \cite{metabaseline} & 63.17 $\pm$ 0.23 & 69.96  $\pm$ 0.21  & \textbf{72.79  $\pm$ 0.19} & 68.62 $\pm$ 0.27 & 73.05  $\pm$ 0.24  & \textbf{73.90  $\pm$ 0.24} \\
      \hline
       1-N-N & 63.28 $\pm$ 0.23 & 67.25  $\pm$ 0.20  & \textbf{69.27  $\pm$ 0.19} & 68.73 $\pm$ 0.26 & 68.05  $\pm$ 0.25  & \textbf{69.82  $\pm$ 0.24} \\
      SVM & 63.41 $\pm$ 0.23 & 70.30  $\pm$ 0.20 & \textbf{72.84  $\pm$ 0.19} & 68.88 $\pm$ 0.25 & 69.26  $\pm$ 0.25  & \textbf{71.28  $\pm$ 0.24}\\
      LR & 63.33  $\pm$ 0.22 & 72.11  $\pm$ 0.20 & \textbf{73.41  $\pm $ 0.19} & 69.15 $\pm$ 0.25 & 74.99  $\pm$ 0.23  & \textbf{75.98  $\pm$ 0.23}\\
      \hline
    \end{tabular}}  
    \caption{\textbf{Choices of the classifiers}. One-shot classification accuracy on \textit{mini}ImageNet and \textit{tiered}ImageNet using different types of classifiers, \textit{i.e.}, 1-N-N, SVM and LR. All methods use the feature extractor from the Meta-Baseline method \cite{metabaseline}.
    %
    }
    \label{tab:classifier}%
  \end{table*}

 \subsection{Performance with Different Classifiers}
 \label{sec: classifier}
 In our main experiments, we classify samples by finding the nearest neighbor among class prototypes.  In this section, we apply another three different types of classifiers: 1-nearest neighbor classifier (1-N-N), Support Vector Machine (SVM), and Logistic Regression (LR).  
 
 Table \ref{tab:classifier} shows the 1-shot performance of different classifiers using our generated features on \textit{mini}ImageNet and \textit{tiered}ImageNet datasets. It shows that the features generated by our VAEs improve the performance of all three classifiers. 
 For example, the 1-shot accuracy on \textit{mini}ImageNet using LR is improved by $8.8\%$ with SVAE and by $10.1\%$ with R-SVAE. The consistent performance improvements show that our generated features can benefit different types of classifiers.
 \subsection{Feature Distribution Analysis}
 
 In Fig. \ref{fig:feats_tsne}, we show the t-SNE representation \cite{Maaten2008VisualizingDU} of different sets of features for three classes from the novel set of \textit{tiered}ImageNet dataset. From left to right, we visualize the distribution of the original support set (a), the query set (b), the features generated by SVAE (c), and the features generated by R-SVAE (d). Note that our methods do not rely on the support features to generate features.
 
 Fig. \ref{fig:feats_tsne}(c) and (d) visualize the effect of our sample selection method. Fig. \ref{fig:feats_tsne}(c) visualizes features generated from our method trained with all available data from the base classes, which consist of 1281 images per class on average. In Fig. \ref{fig:feats_tsne}(d), we train the same model with only 484 representative images per class on average. 
 Our model trained with a representative subset of data generates features that lie closer to the real features, showing the effectiveness of our sample selection method.

 Moreover, we plot the distance distributions between the estimated prototypes and the ground truth prototypes of each class. Specifically, for each class, we first obtain the ground-truth prototype by taking the mean of all the features of the class. Then we calculate the $L_2$ distance between the ground truth prototype and three different prototypes: 1) Baseline: the prototype was estimated using only the support samples. 2) SVAE: the prototype was estimated using the support samples and the generated samples from our SVAE model. 3) R-SVAE: the prototype was estimated using the support samples and the generated samples from our R-SVAE model.

 We sample 2400 tasks from \textit{mini}ImageNet dataset under both 5-way 1-shot and 5-way 5-shot settings. For each task, we obtain five distances, one distance per class. Then we plot the probability density distribution of the distance, shown in Fig. \ref{fig:dist_density}. The probability density is calculated by binning and counting observations and then smoothing them with a Gaussian kernel, namely, Kernel Density Estimation \cite{Chen2017ATO}. As can be seen the Fig., our estimated class prototypes are much closer to the ground truth prototypes, compared to the baseline. 



 \subsection{Sample Visualization}

 In Fig. \ref{fig:img_visualization}, we visualize some representative samples and non-representative samples based on the representativeness probability computed via our method. The samples on the left panel are images with high probabilities. These images mostly contain the main object of the category and are easy to recognize. On the contrary, the samples on the right panel are those with small probabilities. They contain various class-unrelated objects and can lead to noisy features for constructing class prototypes.

\subsection{Performance with Different Semantic Embedding}
We use CLIP features in our main experiments. The performance of our method trained with Word2Vec\cite{Mikolov2013EfficientEO} features are shown in Table \ref{tab:word2vec}. Note that CLIP model is trained with 400M pairs (image and its text title) collected from the web while Word2Vec is trained with only text data. Our model outperforms state-of-the-art methods in both cases.

 \begin{table}[] 
    \centering
  \resizebox{0.45\textwidth}{!}{%
    \begin{tabular}{l|cc}
      \hline
       & 1-shot & 5-shot  \\
      \hline
      Meta-Baseline & 63.17 $\pm$ 0.23 & 79.26 $\pm$ 0.17     \\
      Meta-Baseline + SVAE & 67.39 $\pm$ 0.21 & 79.77 $\pm$ 0.17  \\
      Meta-Baseline + R-SVAE & \textbf{68.03  $\pm$ 0.22} & \textbf{79.93 $\pm$ 0.16} \\
      \hline
    \end{tabular}}  
    \caption{\textbf{Classification accuracy using Word2Vec\cite{Mikolov2013EfficientEO} as the semantic feature extractor.}
    }
    \label{tab:word2vec}%
  \end{table}

\section{Limitations and Discussion} 

   We propose a feature generation method using a conditional VAE model. Here we focus on modeling the distribution of the representative samples rather than the whole data distribution. To accomplish that, we propose a sample selection method to collect a set of strictly representative training samples for training our VAE model. We show that our method brings consistent performance improvements over multiple baselines and achieves state-of-the-art performance on both \textit{mini}ImageNet and \textit{tiered}ImageNet datasets. 
   Our method requires a pre-trained NLP model to obtain the semantic embedding of each class. It might also inherit some potential biases from the textual domain. 
   Note that our method does not aim to generate diverse data with large intra-class variance \cite{JingyiICCV21, metric_learning}. Building a system that can generate both representative and non-representative samples can greatly benefit various downstream computer vision tasks and is an interesting direction to extend our work. 
   
\newcommand{\myheading}[1]{\vspace{1ex}\noindent \textbf{#1}}
\myheading{Acknowledgements.} 
Jingyi Xu is partially supported by a research grant from Zebra Technologies and the SUNY2020 ITSC grant. Hieu Le is funded by Amazon Robotics to attend the conference. We thank Tran Truong, Kien Huynh, and Bento Gonçalves for proofreading the paper.
\FloatBarrier
{\small
\bibliographystyle{ieee_fullname}
\bibliography{egbib,attributes_bib,hieu}
}

\end{document}


\title{Generating Representative Samples for  Few-Shot Classification\\
Supplementary Material}

\author{Jingyi Xu\\
Stony Brook University\\
{\tt\small jingyixu@cs.stonybrook.edu}
\and
Hieu Le\thanks{Work done outside of Amazon}\\
Amazon Robotics\\
{\tt\small ahieu@amazon.com}
}
\maketitle

 \section{Using a deeper network architecture for the decoder}
 \label{sec:vae_underfit}

The decoder of our proposed VAE plays a vital role in our framework as it maps the latent space of the VAE and the semantic embedding to the visual feature embedding space.  We use a network with two fully-connected (FC) layers for the decoder in our main setting. We experiment with a deeper network where we add an FC layer with 4096 hidden units and a LeakyReLU \cite{LReLu} layer to the decoder. Table \ref{tab:deeper_vae} summarizes the results. Using a deeper network degrades the performance of our model under both 1-shot and 5-shot settings for both \textit{mini}ImageNet \cite{Ravi2017OptimizationAA} and \textit{tiered}ImageNet \cite{Ren2018MetaLearningFS_tieredImagenet}. 
 
  \begin{table*}[h!] 
    \centering
  \resizebox{0.65\textwidth}{!}{%
    \begin{tabular}{l|cc|cc}
      \hline
        & \multicolumn{2}{c|}{\textit{mini}ImageNet} & \multicolumn{2}{c}{\textit{tiered}ImageNet}
           \\
          Decoder & 1-shot & 5-shot &  1-shot & 5-shot \\
      \hline
       2-FC Layers (Main paper) & 72.79 $\pm$ 0.19 & 80.70  $\pm$ 0.16 & {74.21  $\pm$ 0.24} & 84.17 $\pm$ 0.18  \\
      3-FC Layers & 71.68 $\pm$ 0.20 & 80.32  $\pm$ 0.16 & {73.84  $\pm$ 0.24} & 84.05 $\pm$ 0.25 \\
      \hline
    \end{tabular}}  
    \caption{\textbf{Few-shot classification performance of our method using different network architecture for the decoder}. In our main setting, we use as our decoder a network with two fully-connected (FC) layers. ``3-FC Layers'' denotes the setting where we add an FC layer with 4096 hidden units and a LeakyReLU layer to the decoder. The performance degrades for both 1-shot and 5-shot settings with a deeper network.
    %
    }
    \label{tab:deeper_vae}%
  \end{table*}
  
\section{Sample visualization}
 \label{sec:visualization}

In Figure \ref{fig:img_visualization}, we provide additional visualization of some representative samples
and non-representative samples based on the representativeness probability computed via our method. The samples on
the left panel are images with high probabilities. These images mostly contain the main object of the category and are
easy to recognize. On the contrary, the samples on the right
panel are those with small probabilities. They contain various class-unrelated objects and can lead to noisy features
for constructing class prototypes.

  \def\subboxsize{0.48\textwidth}
 \begin{figure*}[ht!]
 \centering
\hspace*{-0.1cm}\includegraphics[width=\linewidth]{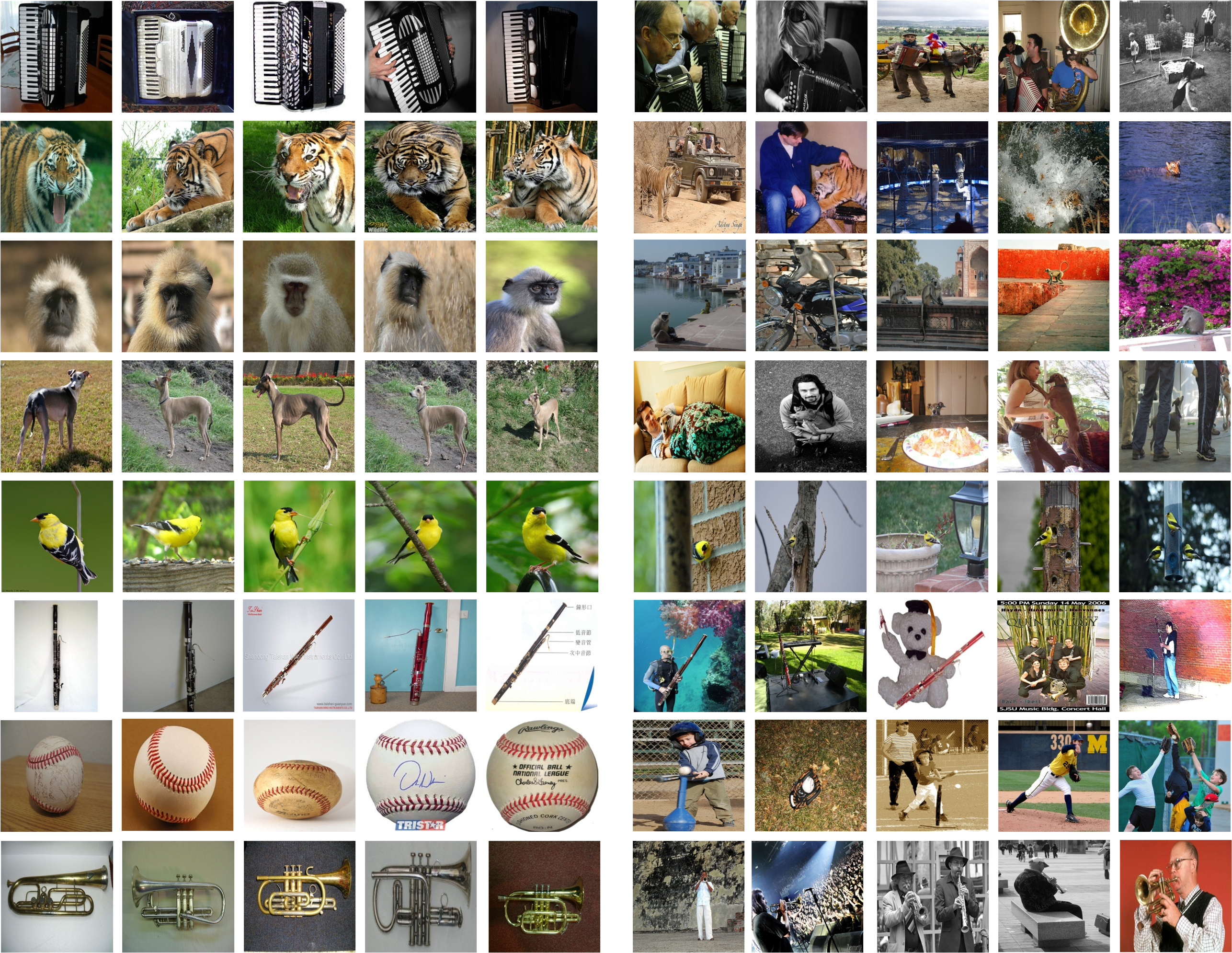}
\makebox[\subboxsize]{Representative }
\makebox[\subboxsize]{Non-representative }
 \caption{ \textbf{Examples of representative samples (left) and non-representative samples (right)}. We visualize 5 images with high probabilities and 5 images with small probabilities computed via our proposed method for 8 classes from \textit{tiered}ImageNet dataset.
    }

    \label{fig:img_visualization}
\end{figure*}

 \section{Analysis of base class prototypes with the proposed sample selection method}
 
 Our feature selection method filters out non-representative samples from the base classes before training our VAE model. As shown in the main paper - Figure 3, our method performs better as the representativeness threshold increases.  A higher threshold means we select samples that are more representative, resulting in a less amount of training data points. The second column of  Table \ref{tab:improvement} shows the percentages of selected samples via our method for 10 classes of the \textit{mini}ImageNet dataset. Here we use a common threshold of 0.9 for all classes. Note that the average number of available training data for each class is 600. Our method achieves state-of-the-art few-shot classification performance when using only a small fraction of these training data. For example, we only use 42 images from class ``Wok'' and 91 from class ``Jellyfish'' to train our VAE model.
 
We observe a correlation between the numbers of selected samples and the distances between the estimated prototypes and the ground truth prototypes of the base classes. We first estimate a \textit{ground truth} prototype for each base class using all available features. The prototype is computed as the mean of all features. We then train a baseline VAE model using all available training data from the base classes to compare with our model trained using only the representative data. For each class, we obtain two prototypes using the generated features from the baseline model and our VAE model. $\mathcal{D}_{\textrm{all-data}}$ denotes the distances between the prototypes estimated using the baseline model and the ground truth prototypes. $\mathcal{D}_{\textrm{selected-data}}$ denotes the distances for our proposed model. As can be seen from the last column of Table \ref{tab:improvement}, our VAE model trained with only representative samples approximates better the ground truth prototypes. Moreover, the improvements are more pronounced for classes with small amounts of training samples.

 \label{sec:distribution_analysis}
\begin{table*}[h!] 
    \centering
  \resizebox{0.7\textwidth}{!}{%
    \begin{tabular}{l|c|c}
      \hline
    Class name    &  \% selected data   & $\mathcal{D}_{\textrm{all-data}}\rightarrow\mathcal{D}_{\textrm{selected-data}} (\downarrow \textrm{improvement})$ \\
      \hline
      Wok & 7.0\% & 0.68 $\rightarrow$ 0.51 ($\downarrow$ 0.17) \\
      Parallel bars & 4.3\% &  0.67 $\rightarrow$ 0.52  ($\downarrow$ 0.15) \\
      Green Mamba & 6.7\% & 0.68 $\rightarrow$ 0.54 ($\downarrow$ 0.14) \\
      Bolete & 6.0\% & 0.61 $\rightarrow$ 0.50  ($\downarrow$ 0.11) \\
      Boxer & 5.8\% & 0.76 $\rightarrow$ 0.64 ($\downarrow$ 0.12) \\
      \hline
      Jellyfish & 15.2\% & 0.66 $\rightarrow$ 0.62 ($\downarrow$ 0.04) \\
      Dugong & 15.7\% & 0.69 $\rightarrow$ 0.64 ($\downarrow$ 0.05) \\
      Spider web & 17.3\% & 0.64 $\rightarrow$ 0.59 ($\downarrow$ 0.05) \\
      Snorkel & 13.7\% & 0.68 $\rightarrow$ 0.64 ($\downarrow$ 0.04) \\
      Hair Slide & 13.5\% & 0.50 $\rightarrow$ 0.44 ($\downarrow$ 0.06) \\
      \hline
    \end{tabular}}  
    \caption{\textbf{Percentages of representative samples.} We show the percentages of representative samples for 10 classes of the \textit{tiered}ImageNet dataset, selected via our sample selection method. The VAE model trained only with these representative data estimates better the ground truth prototypes of the base classes.$\mathcal{D}_{\textrm{all-data}}$ denotes the distances between the prototypes estimated using the VAE model trained with all data and the ground truth prototypes. $\mathcal{D}_{\textrm{selected-data}}$ denotes the distances between the prototypes estimated using the VAE model trained with only the selected data and the ground truth prototypes.
    }
    \label{tab:improvement}%
  \end{table*}
 
 \section{1-shot classification accuracy with different support images}
 \label{sec:one-shot}

\begin{figure}[h]
\begin{center}
\includegraphics[width=0.5\linewidth]{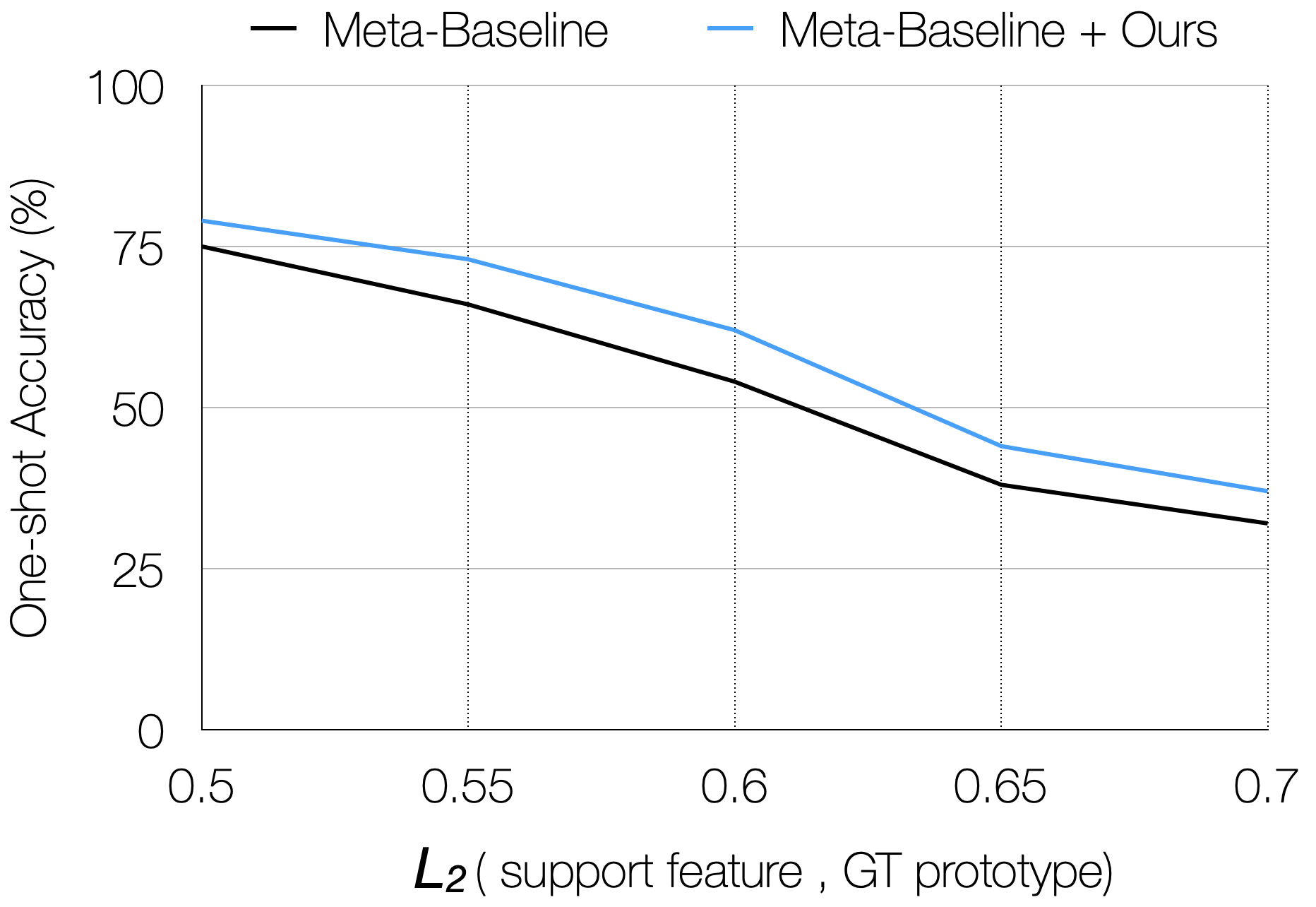}
\end{center}
\caption{
\textbf{1-shot, 5-way classification accuracy with different support images. }
FSL methods heavily depend on the representativeness of the support samples. The figure shows the 1-shot, 5-way accuracy of the Meta-Baseline method \cite{metabaseline} and our method.  Here we fix the 5 classes used for evaluation and experiment with 5 different support images for one of the 5 classes. These support images have different $L_{2}$ distances to the mean feature of the class (\textit{i.e.}, ground truth prototype), ranging from 0.5 to 0.7. A smaller value means the support feature is more representative. As can be seen, the performance of both Meta-Baseline and our method decreases dramatically when the representativeness of the support sample decreases. 
}
\label{fig:one-shot}
\end{figure}

We observe that performance of few-shot learning methods heavily depends on the representativeness of the support samples. For example, Figure \ref{fig:one-shot} shows the 5-way 1-shot accuracy of the Meta-Baseline method \cite{metabaseline} and our method. Here we fix the 5 classes used for evaluation and experiment with 5 different support images for one of the 5 classes. These support images have different $L_{2}$ distances to the mean feature of the class (\textit{i.e.}, ground truth prototype), ranging from 0.5 to 0.7. A smaller value means the support feature is more representative. As can be seen, the performance of both Meta-Baseline and our method decreases dramatically when the representativeness of the support sample decreases. 

 \section{Zero-shot Performance} 
 
We show the effectiveness of our generated features without any few-shot sample given. In this case, the prototype is obtained by only taking the mean of the features generated by our VAE models. As shown in Table \ref{tab:zero-shot}, using our 0-shot prototypes outperforms the 1-shot prototypes estimated from the real sample. 
\begin{table}[!h] 
    \centering
  \resizebox{0.45\textwidth}{!}{%
    \begin{tabular}{l|cc}
      \hline
       & Meta-Baseline \cite{metabaseline} & ProtoNet \cite{Ye2020FewShotLV}  \\
      \hline
      \hline
      1-shot & 63.17 $\pm$ 0.23 & 62.39     \\
      0-shot with SVAE & 66.42 $\pm$ 0.22 & 65.47  $\pm$ 0.41  \\
      0-shot with R-SVAE & \textbf{66.76  $\pm$ 0.21} & \textbf{69.23 $\pm$ 0.39} \\
      \hline
    \end{tabular}}  
\caption{\textbf{Zero-shot classification accuracy.} We construct class prototypes using only generated features from SVAE and R-SVAE. ``1-shot'' indicates the performance of the baseline methods in the 1-shot setting using support features.
    }
    \label{tab:zero-shot}%
    \end{table}

\FloatBarrier
\section{Performance on CIFAR-FS and FC-100}
In Table \ref{tab:cifar-fs}, we provide the performance of our method on two additional FSL datasets - CIFAR-FS and FC-100. On these both datasets, our method improves the Meta-Baseline method by large margins.
\begin{table}[!h] 
    \centering
  \resizebox{0.45\textwidth}{!}{%
    \begin{tabular}{l|c|cc}
      \hline
       & Dataset & 1-shot & 5-shot  \\
      \hline
       Meta-Baseline &  & 64.96 $\pm$ 0.51 & 75.85 $\pm$ 0.40 \\
       Meta-Baseline + SVAE & CIFAR-FS  & 72.07 $\pm$ 0.45 & 77.18  $\pm$ 0.39  \\
       Meta-Baseline + R-SVAE &   & \textbf{73.25 $\pm$ 0.44} & \textbf{78.89 $\pm$ 0.37} \\
      \hline
       Meta-Baseline & & 41.31 $\pm$ 0.42 & 51.84 $\pm$ 0.40 \\
       Meta-Baseline + SVAE & FC-100 & 45.65 $\pm$ 0.40 & 54.37 $\pm$ 0.40  \\
       Meta-Baseline + R-SVAE & & \textbf{45.75 $\pm$ 0.40} & \textbf{54.44 $\pm$ 0.40} \\
      \hline
    \end{tabular}}  
    \caption{1-shot and 5-shot classification accuracy on CIFAR-FS and FC-100.}
    \label{tab:cifar-fs}
  \end{table}

\section{Comparison with methods using semantic information.} In Table \ref{tab:semantics}, we compare our method with
the FSL methods using semantics information including TriNet \cite{closer-look}, CFA \cite{Hu2019WeaklysupervisedCF}, FSLKT \cite{Peng2019FewShotIR},  and AM3 \cite{Xing2019AdaptiveCF_AM3}. 
We did not include the results of ProtoComNet since it is under transductive setting. 

\begin{table}[!h] 
    \centering
  \resizebox{0.45\textwidth}{!}{%
    \begin{tabular}{l|c|cc}
      \hline
       & backbone & 1-shot & 5-shot  \\
      \hline
      TriNet  & ResNet18 & 58.12 $\pm$ 1.37 & 76.92 $\pm$ 0.69 \\
      CFA & ResNet18 & 58.5 $\pm$ 0.8 & 76.6 $\pm$ 0.6 \\
      FSLKT  & ConvNet(128F) & 64.42 $\pm$ 0.72 & 74.16 $\pm$ 0.56  \\
      AM3  & ResNet12 & 65.30 $\pm$ 0.49 & 78.10 $\pm$ 0.36  \\
      Ours & ResNet12 & \textbf{74.84  $\pm$ 0.23} & \textbf{83.28 $\pm$ 0.40} \\
      \hline
    \end{tabular}}  

    \caption{Comparison to prior semantic-based methods on \textit{mini}ImageNet.}
    \label{tab:semantics}%

  \end{table}

\section{Ablation study for the sample selection method.} 
We compare our sample selection method based on Gaussian distribution with other methods including herding \cite{Rebuffi2017iCaRLIC} and K-means selection \cite{Chang2021OnTI} in Table \ref{tab:cluster}. The experiment is conducted on the \textit{mini}ImageNet dataset.
\begin{table}[!h] 

    \centering
  \resizebox{0.4\textwidth}{!}{%
    \begin{tabular}{l|cc}
      \hline
       & 1-shot & 5-shot  \\
      \hline
      Baseline & 69.96 $\pm$ 0.21 & 79.92 $\pm$ 0.16 \\
      Herding & 72.14 $\pm$ 0.20 & 80.48  $\pm$ 0.16  \\
      K-means selection & {72.31  $\pm$ 0.20} & {80.55 $\pm$ 0.16} \\
      Ours (Gaussian) & \textbf{72.79  $\pm$ 0.19} & \textbf{80.70 $\pm$ 0.16} \\
      \hline
    \end{tabular}}  
        \caption{1-shot and 5-shot classification accuracy on \textit{mini}ImageNet using different clustering methods.}
    \label{tab:cluster}

  \end{table}

\section{Comparison with augmentation-based methods} 

We show the results of our method in comparison with state-of-the-art augmentation-based methods in Table \ref{tab:aug}.  These methods include MetaGAN \cite{metagan}, AFHN \cite{adversarial2020kai}, Delta-Encoder \cite{delta-encoder}, IDeMe-Net\cite{Chen2019ImageDM}, MABAS \cite{kim2020_mabas} and DC \cite{Yang2021FreeLF}.
\begin{table}[!h] 
    \centering
  \resizebox{0.45\textwidth}{!}{%
    \begin{tabular}{l|c|cc}
      \hline
       & backbone & 1-shot & 5-shot  \\
      \hline
      MetaGAN & ResNet18 & 52.71 $\pm$ 0.64 & 68.63 $\pm$ 0.67 \\
      AFHN  & ResNet18 & 62.38 $\pm$ 0.72 & 78.16 $\pm$ 0.56 \\
      Delta-Encoder & ResNet18 & 59.90 & 69.70  \\
      IDeMe-Net & ResNet10 & 59.14 $\pm$ 0.86 & 74.63 $\pm$ 0.74 \\ 
      MABAS & ResNet12 & 65.08$ \pm$ 0.86 & 82.70 $\pm$ 0.54  \\
      DC & WRN28 & 68.57 $\pm$ 0.55 & 82.88 $\pm$ 0.42 \\ 
      Ours & ResNet12 & \textbf{74.84  $\pm$ 0.23} & \textbf{83.28 $\pm$ 0.40} \\
      \hline
    \end{tabular}}  

    \caption{Comparison to prior augmentation-based methods on \textit{mini}ImageNet.}
    \label{tab:aug}%

  \end{table}

\newpage
\FloatBarrier
{\small
\bibliographystyle{ieee_fullname}
\bibliography{egbib,attributes_bib}
}